%% file: liu2022ral.tex
\documentclass[letterpaper, 10pt, conference]{ieeeconf}
\usepackage{graphicx}
\IEEEoverridecommandlockouts
\usepackage{float}
\usepackage{rotating}
\usepackage{subcaption}
\usepackage{psfrag}
\usepackage{amsmath,amssymb,amsfonts,bbm,nicefrac,mathtools,lipsum}
\usepackage{multirow}
\usepackage{latexsym}
\usepackage{caption}
\usepackage{balance}
\usepackage{colortbl}
\usepackage[table]{xcolor}
\usepackage[super]{nth}
\usepackage{fancyhdr}
\usepackage{epstopdf}
\usepackage{float}
\usepackage[hidelinks]{hyperref}
\usepackage{booktabs,tabularx}
\usepackage{makecell}
\usepackage{dirtytalk}
\usepackage[noend]{algpseudocode}
\usepackage{siunitx}
\usepackage{academicons}

\makeatletter
\usepackage{xspace}
\DeclareRobustCommand\onedot{\futurelet\@let@token\@onedot}
\def\@onedot{\ifx\@let@token.\else.\null\fi\xspace}
 
\def\ie{i.e\onedot}

\def\etal{{et al}\onedot}
\def\etalcite#1{\etal~\cite{#1}}
\makeatother

\usepackage{soul}
\usepackage[ruled]{algorithm2e}
\SetAlCapSkip{10em}
\SetAlgoSkip{bigskip}
\usepackage[capitalize]{cleveref}

\newcommand{\sw}[1]{\begin{sideways}#1\end{sideways}}

\usepackage[acronyms, shortcuts]{glossaries}
\glsdisablehyper

\newacronym{ioc}{IOC}{inverse optimal control}
\newacronym{lqr}{LQR}{linear-quadratic regulator}
\newacronym{kkt}{KKT}{Karush–Kuhn–Tucker}
\newacronym{irl}{IRL}{inverse reinforcement learning}
\newacronym{mle}{MLE}{maximum likelihood estimation}
\newacronym{sem}{SEM}{standard error of the mean}
\newacronym{mpgp}{MPGP}{model-predictive game play}
\newacronym[longplural=open-loop Nash equilibria,plural=OLNE]{olne}{OLNE}{open-loop Nash equilibrium}
\newacronym{licq}{LICQ}{linear independence constraint qualification}
\newacronym{mpc}{MPC}{model-predictive control}
\newacronym{mpec}{MPEC}{mathematical program with equilibrium constraints}
\newacronym[longplural={partially observable Markov decision processes}]{pomdp}{POMDP}{partially observable Markov decision process}
\newacronym{nep}{NEP}{Nash equilibrium problem}
\newacronym{gnep}{GNEP}{generalized Nash equilibrium problem}
\newacronym{gne}{GNE}{generalized Nash equilibrium}
\newacronym{svo}{SVO}{social value orientation}
\newacronym{ukf}{UKF}{unscented Kalman filter}
\newacronym{ibr}{IBR}{iterated best response}
\newacronym{awgn}{AWGN}{additive white Gaussian noise}
\newacronym{iqr}{IQR}{interquartile range}
\newacronym{mcp}{MCP}{Mixed Complementarity Problem}
\newacronym{ift}{IFT}{implicit function theorem}
\newacronym{nn}{NN}{neural network}
\newcommand{\given}{\mid}
\newcommand{\mbb}{\mathbb}
\newcommand{\mc}{\mathcal}

\newcommand{\R}{\mbb{R}}

\newcommand{\numplayers}{N}

\newcommand{\dynamics}{f}

\newcommand{\state}{x}

\newcommand{\control}{u}

\newcommand{\capcontrol}{U}
\newcommand{\capbcontrol}{\mathbf{\capcontrol}}
\newcommand{\capstate}{X}
\newcommand{\capbstate}{\mathbf{\capstate}}
\newcommand{\bstate}{\mathbf{\state}}
\newcommand{\bcontrol}{\mathbf{\control}}
\newcommand{\xdim}{n}

\newcommand{\horizon}{T}
\newcommand{\cost}{J}

\newcommand{\observation}{y}
\newcommand{\bobservation}{\mathbf{\observation}}
\newcommand{\capobservation}{Y}
\newcommand{\capbobservation}{\mathbf{\capobservation}}

\newcommand{\game}{\Gamma}

\newcommand{\pmcp}{\Psi}

\newcommand{\example}[1]%
{
\vspace{0.15cm}
\noindent \textit{\textbf{Running example:} #1}
\vspace{0.15cm}
}

\newtheorem{definition}{Definition}

\newcommand{\revised}[1]{{\leavevmode\color{black}#1}}
\newcommand{\revisedInline}[1]{\textcolor{black}{#1}}

\graphicspath{{./Images/}}

\begin{document}

\title{\LARGE \bf Learning to Play Trajectory Games\\ Against Opponents with Unknown Objectives}

\author{Xinjie Liu$^{\star}$, Lasse Peters$^{\star}$, and Javier Alonso-Mora
\thanks{
This work is funded in part by the European Union (ERC, INTERACT, 101041863). Views and opinions expressed are however those of the author(s) only and do not necessarily reflect those of the European Union or the European Research Council Executive Agency. Neither the European Union nor the granting authority can be held responsible for them.
All authors are with the Department of Cognitive Robotics (CoR), Delft University
of Technology, 2628 CD Delft, Netherlands (email: \tt x.liu-47@student.tudelft.nl; l.peters@tudelft.nl; j.alonsomora@tudelft.nl).
\smallskip
}
}
\maketitle
\thispagestyle{empty}
\pagestyle{empty}

\def\thefootnote{$\star$}\footnotetext{Equal contribution (Corresponding author: Xinjie Liu).}\def\thefootnote{\arabic{footnote}}
\begin{abstract}

Many autonomous agents, such as intelligent vehicles, are inherently required to interact with one another.
Game theory provides a natural mathematical tool for robot motion planning in such interactive settings.
However, tractable algorithms for such problems usually rely on a strong assumption, namely that the objectives of all players in the scene are known.
To make such tools applicable for ego-centric planning with only local information, we propose an adaptive model-predictive game solver, which jointly infers other players’ objectives online and computes a corresponding \ac{gne} strategy. %
The adaptivity of our approach is enabled by a differentiable trajectory game solver whose gradient signal is used for \ac{mle} of opponents' objectives.
This differentiability of our pipeline facilitates direct integration with other differentiable elements, such as \acp{nn}.
Furthermore, in contrast to existing solvers for cost inference in games, our method handles not only partial state observations but also general inequality constraints.
In two simulated traffic scenarios, we find superior performance of our approach over both existing game-theoretic methods and non-game-theoretic \ac{mpc} approaches.
We also demonstrate our approach's real-time planning capabilities and robustness in two hardware experiments.

\end{abstract}

\vspace{0.1cm}

\begin{keywords}
Trajectory games, multi-robot systems, integrated planning and learning, human-aware motion planning.
\end{keywords}
\vspace{-1.5em}

\section{Introduction}
\label{sec:intro}

Many robot planning problems, such as robot navigation in a crowded environment, involve rich interactions with other agents. Classic \say{predict-then-plan} frameworks neglect the fact that other agents in the scene are responsive to the ego-agent's actions. This simplification can result in inefficient or even unsafe behavior~\cite{trautman2010unfreezing}. Dynamic game theory explicitly models the interactions as coupled trajectory optimization problems from a multi-agent perspective. A noncooperative equilibrium solution of this game-theoretic model then provides strategies for all players that account for the strategic coupling of plans.
Beyond that, general constraints between players, such as collision avoidance, can also be handled explicitly.
All of these features render game-theoretic reasoning an attractive approach to interactive motion planning.

\begin{figure}[t]
  \centering
  \includegraphics[width=0.8\linewidth, trim={0 1.25em 0 0},clip]{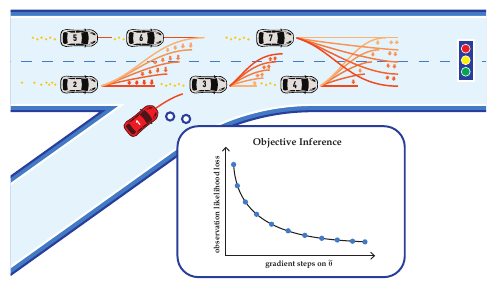}
  \caption{
  An ego-agent (red) merging onto a busy road populated by six surrounding vehicles whose preferences for travel velocity and lane are initially unknown.
  Our approach adapts the ego agent's strategy by inferring opponents' intention parameters $\tilde{\theta}$ from partial state observations.
  }
  \label{fig:motivation}
\end{figure}

In order to apply game-theoretic methods for interactive motion planning from an \emph{ego-centric} rather than \emph{omniscient} perspective, such methods must be capable of operating only based on local information. For instance, in driving scenarios as shown in~\cref{fig:motivation}, the red ego-vehicle may only have partial-state observations of the surrounding vehicles and incomplete knowledge of their objectives due to unknown preferences for travel velocity, target lane, or driving style. Since vanilla game-theoretic methods require an objective model of \emph{all} players~\cite{fridovich2020efficient,cleac2022algames}, this requirement constitutes a key obstacle in applying such techniques for autonomous strategic decision-making. %

To address this challenge, we introduce our main contribution: a model-predictive game solver, which adapts to unknown opponents' objectives and solves for \acf{gne} strategies. %
The adaptivity of our approach is enabled by a differentiable trajectory game solver whose gradient signal is used for \ac{mle} of opponents' objectives.

\revised{
We perform thorough experiments in simulation and on hardware to support the following three key claims: our solver
(\romannumeral 1)~outperforms both game-theoretic and non-game-theoretic baselines in highly interactive scenarios,
(\romannumeral 2)~can be combined with other differentiable components such as \acp{nn}, and
(\romannumeral 3)~is fast and robust enough for real-time planning on a hardware platform.
}

\section{Related Work}
\label{sec: related_work}

To put our contribution into context, this section discusses \revised{four} main bodies of related work. First, we discuss works on trajectory games which assume access to the objectives of all players in the scene.
Then, we introduce works on inverse dynamic games that infer unknown objectives from data.
\revised{Thereafter, we also relate our work to non-game-theoretic interaction-aware planning-techniques.} 
\revised{Finally}, we survey recent advances in differentiable optimization, which provide the underpinning for our proposed differentiable game solver.

\subsection{\numplayers-Player General-Sum Dynamic Games}
Dynamic games are well-studied in the literature~\cite{Basar1998games}.
In robotics, a particular focus is on multi-player general-sum games in which players may have differing yet non-adversarial objectives, and states and inputs are continuous.

Various equilibrium concepts exist in dynamic games.
The Stackelberg equilibrium concept~\cite{liniger2019noncooperative} assumes a \say{leader-follower} hierarchy, while the \ac{nep}~\cite{fridovich2020efficient, liniger2019noncooperative}
does not presume such a hierarchy.
Within the scope of \ac{nep}, there exist open-loop \acp{nep}~\cite{cleac2022algames} and feedback \acp{nep}~\cite{fridovich2020efficient, laine2021computation}.
We refer the readers to~\cite{Basar1998games} for more details about the difference between the concepts.
When shared constraints exist between players, such as collision avoidance constraints, one player's feasible set may depend on other players' decisions.
In that case, the problem becomes a \ac{gnep}~\cite{facchinei2010generalized}.
In this work, we focus on \acp{gnep} under an open-loop information pattern which we solve by converting to an equivalent \ac{mcp}~\cite{mcp_ref}.

\subsection{Inverse Games}\label{inverse survey}

There are three main paradigms for solving inverse games: (\romannumeral 1) Bayesian inference, (\romannumeral 2) minimization of \ac{kkt} residuals, and (\romannumeral 3) equilibrium-constrained maximum-likelihood estimation.
In type~(\romannumeral 1) methods, Le~Cleac'h~\etalcite{le2021lucidgames} employ an Unscented Kalman Filter (UKF).
This sigma-point sampling scheme drastically reduces the sampling complexity compared to vanilla particle filtering.
However, a UKF is only applicable for uni-modal distributions, and extra care needs to be taken when uncertainty is multi-modal, e.g., due to multiple Nash equilibria.
Type~(\romannumeral 2) methods require \textit{full} demonstration trajectories, i.e., including noise-free states and inputs, to cast the $N$-player inverse game as $N$ independent unconstrained optimization problems~\cite{awasthi2020inverse, ROTHFU201714909}. However, they assume full constraint satisfaction at the demonstration and have limited scalability with noisy data~\cite{peters2021inferring}. 
The type~(\romannumeral 3) methods use \ac{kkt} conditions of an \ac{olne} as constraints to formulate a constrained optimization problem~\cite{peters2021inferring}.
This type of method finds the same solution as type (\romannumeral 2) methods in the noise-free cases but can additionally handle partial and noisy state observations.
However, encoding the equilibrium constraints is challenging, as it typically yields a non-convex problem, even in relatively simple linear-quadratic game settings.
This challenge is even more pronounced when considering inequality constraints of the observed game, as this results in complementarity constraints in the inverse problem.

Our solution approach also matches the observed trajectory data in an \ac{mle} framework.
In contrast to all methods above, we do so by making a \ac{gne} solver differentiable.
This approach yields two important benefits over existing methods: (\romannumeral 1) general (coupled) inequality constraints can be handled explicitly, and (\romannumeral 2) the entire pipeline supports direct integration with other differentiable elements, such as \acp{nn}.
\revised{
This latter benefit is a key motivation for our approach that is not enabled by the formulations in \cite{le2021lucidgames} and \cite{peters2021inferring}.
}

Note that Geiger~\etalcite{geiger2021learning} explore a similar differentiable pipeline for inference of game parameters.
In contrast to their work, however, our method is not limited to the special class of potential games and applies to general \acp{gnep}.

\revised{
\subsection{Non-Game-Theoretic Interaction Models}
Besides game-theoretic methods, two categories of interaction-aware decision-making techniques have been studied extensively in the context of collision avoidance and autonomous driving:~(\romannumeral 1) approaches that learn a navigation policy for the ego-agent directly without explicitly modeling the responses of others~\cite{everett2018motion,brito2021go,tolani2021visual}, and (\romannumeral 2) techniques that explicitly predict the opponents' actions to inform the ego-agent's decisions~\cite{kretzschmar2016socially,schmerling2018multimodal,rhinehart2019precog,roh2021multimodal,sun2021move}.
This latter category may be further split by the granularity of coupling between the ego-agent's decision-making process and the predictions of others.
In the simplest case, prediction depends only upon the current physical state of other agents%
~\cite{scholler2020constant}.
More advanced interaction models condition the behavior prediction on additional information such as the interaction history~\cite{kretzschmar2016socially}, the ego-agent's goal~\cite{rhinehart2019precog,roh2021multimodal}, or even the ego-agent's future trajectory~\cite{schmerling2018multimodal,sun2021move}.

Our approach is most closely related to this latter body of work: by solving a trajectory game, our method captures the interdependence of future decisions of all agents; and by additionally inferring the objectives of others, predictions are conditioned on the interaction history.
However, a key difference of our method is that it explicitly models others as rational agents unilaterally optimizing their own cost.
This assumption provides additional structure and offers a level of interpretability of the inferred behavior.
}

\subsection{Differentiable Optimization}
\label{optnet}

Our work is enabled by differentiating through a \ac{gne} solver.
Several works have explored the idea of propagating gradient information through optimization algorithms \cite{ralph1995directional,amos2017optnet,agrawal2019differentiable}, enabling more expressive neural architectures.
However, these works focus on optimization problems and thus only apply to special cases of games, such as potential games studied by~Geiger~\etalcite{geiger2021learning}.
By contrast, differentiating through a~\ac{gnep} involves~$\numplayers$ \emph{coupled} optimization problems.
We address this challenge in section~\ref{differentiability}.

\section{Preliminaries}
\label{sec:preliminaries}

This section introduces two key concepts underpinning our work: forward and inverse dynamic games.
In \emph{forward} games, the objectives of players are known, and the task is to find players' strategies.
By contrast, \emph{inverse} games take (partial) observations of strategies as inputs to recover initially \emph{unknown objectives}.
In \cref{sec:approach}, we combine these two approaches into an adaptive solver that computes forward game solutions while estimating player objectives.

\subsection{General-Sum Trajectory Games}
\label{sec:forward-game}

Consider an~$\numplayers$-player discrete-time general-sum trajectory game with horizon of~$\horizon$.
In this setting, each player~$i$ has a control input~$\control_t^i \in \R^{m^i}$ which they may use to influence the their state~$\state_t^i \in \R^{\xdim^i}$ at each discrete time~$t \in [\horizon]$.
In this work, we assume that the evolution of each player's state is characterized by an individual dynamical system $\state_{t+1}^i = \dynamics^i(\state_t^i, \control_t^i)$.
For brevity throughout the remainder of the paper, we shall use boldface to indicate aggregation over players and capitalization for aggregation over time, e.g.,~$\bstate_t := (\state_t^1, \dots, \state_t^{\numplayers})$,~$\capcontrol^i := (\control_1^i, \dots, \control_{\horizon}^i)$, $\capbstate:= (\bstate_1, \ldots, \bstate_\horizon)$.
With a joint trajectory starting at a given initial state~$\hat{\bstate}_1~:=~(\hat{\state}_1^1, \dots, \hat{\state}_1^\numplayers)$, each player seeks to find a control sequence~$\capcontrol^i$ to minimize their own cost function~$\cost^i(\capbstate, \capcontrol^i; \theta^i)$, which depends upon the joint state trajectory~$\capbstate$ as well as the player's control input sequence~$\capcontrol^i$ and, additionally, takes in a parameter vector~$\theta^i$.\footnote{The role of the parameters will become clear later in the paper when we move on to \emph{inverse} dynamic games.}
Each player must additionally consider private inequality constraints~${^pg^i}(\capstate^i, \capcontrol^i)~\geq~0$ as well as shared constraints~$^{s}g(\capbstate, \capbcontrol) \geq 0$.
This latter type of constraint is characterized by the fact that all players have a shared responsibility to satisfy it, with a common example being collision avoidance constraints between players.
In summary, this noncooperative trajectory game can be cast as a tuple of~$\numplayers$ coupled trajectory optimization problems:
~\begin{equation}
\label{forward_game}
\forall i \in [\numplayers]
\left\{
\begin{split}
        \mathop{\min}\limits_{\capstate^i, \capcontrol^i} \hspace{0.2cm} & {\cost^i(\capbstate, \capcontrol^i; \theta^i)} & \\
        \text{s.t.} \hspace{0.2cm} & \state_{t+1}^i = \dynamics^i(\state_t^i, \control_t^i), \forall t \in [T-1] \\
        & \state_1^i = \hat{\state}_1^i\\
        & ^{p}g^i(\capstate^i, \capcontrol^i) \geq 0 &\\
        & ^{s}g(\capbstate, \capbcontrol) \geq 0. & \\
\end{split}
\right.
\end{equation}

Note that each player's feasible set in this problem may depend upon the decision variables of others, which makes it a \acrshort{gnep} rather than a standard \ac{nep}~\cite{facchinei2010generalized}.

A solution of this problem is a tuple of \ac{gne} strategies~$\capbcontrol^* := (\capcontrol^{1*}, \dots, \capcontrol^{\numplayers*})$ that satisfies the inequalities~$\cost^i(\capbstate^*, \capcontrol^{i*}; \theta^i) \leq \cost^i((\capstate^i, \capbstate^{\neg i *}), \capcontrol^{i}; \theta^i)$ for any feasible deviation $(\capstate^i, \capcontrol^{i})$ of any player $i$, with $\capbstate^{\neg i}$ denoting all but player~$i$'s states.
Since identifying a global \ac{gne} is generally intractable, we require these conditions only to hold locally.
At a local \ac{gne}, then, no player has a unilateral incentive to deviate \textit{locally} in feasible directions to reduce their cost.

\textbf{Running example:}\label{running example} We introduce a simple running example\footnote{
Our final evaluation in \cref{sec:expriments} features denser interaction such as the 7-player ramp-merging scenario shown in \cref{fig:motivation}.} which we shall use throughout the presentation to concretize the key concepts.
Consider a tracking game played between~$\numplayers = 2$ players. Let each agent's dynamics be characterized by those of a planar double-integrator, where states~$x^i_t = (p_{x, t}^i, p_{y, t}^i, v_{x, t}^i, v_{y, t}^i)$ are position and velocity, and control inputs~$u^i_t = (a_{x, t}^i, a_{y, t}^i)$ are acceleration in horizontal and vertical axes in a Cartesian frame.
We define the game's state as the concatenation of the two players' individual states~$\bstate_t := (x_t^1, x_t^2)$.
Each player's objective is characterized by an individual cost
~\begin{multline}
    \label{cost structure}
    \cost^i = \sum_{t = 1}^{T-1} \|p_{t+1}^i - p^i_{\text{goal}}\|_2^2 + 0.1 \| u_t^i \|_2^2
    \\+ 50 \max(0, d_{\min} - \|p_{t+1}^i - p_{t+1}^{-i}\|_2)^3,
\end{multline}

\noindent where we set $p_{\text{goal}}^1 = p_t^2$ so that player~$1$, the tracking robot, is tasked to track player~$2$, the target robot.
Player~$2$ has a fixed goal point~$p_{\text{goal}}^2$.
Both agents wish to get to their goal position efficiently while avoiding proximity beyond a minimal distance~$d_{\text{min}}$.
Players also have shared collision avoidance constraints~${^sg_{t+1}}(\bstate_{t+1},\bcontrol_{t+1}) = \|p_{t+1}^1 - p_{t+1}^2\|_2 - d_{\min} \geq 0, \forall t \in [T-1]$ and private bounds on state and controls ~$^{p}g^i(\capstate^i, \capcontrol^i)$. Agents need to negotiate and find an underlying equilibrium strategy in this noncooperative game, as no one wants to deviate from the direct path to their goal.

\subsection{Inverse Games}
\label{sec:inverse-game}

We now switch context to the \emph{inverse} dynamic game setting.
Let $\theta := (\hat{\bstate}_1, \theta^2, ..., \theta^{\numplayers})$ denote the aggregated tuple of parameters initially unknown to the ego-agent with index~1. Note that we explicitly infer the initial state of a game~$\hat{\bstate}_1$ to account for the potential sensing noise and partial state observations.
To model the inference task over these parameters, we assume that the ego-agent observes behavior originating from an unknown Nash game~$\game(\theta) := (\hat{\bstate}_1, {^sg}, \{\dynamics^i, ^pg^i, J^i(\cdot;\theta^i)\}_{i \in [\numplayers]})$, with objective functions and constraints parameterized by initially unknown values~$\theta^i$ and $\hat{\bstate}_1$, respectively.

Similar to the existing method~\cite{peters2021inferring}, we employ an \ac{mle} formulation to allow observations to be \emph{partial} and \emph{noise-corrupted}.
In contrast to that method, however, we also allow for inequality constraints in the hidden game.
That is, we propose to solve

\begin{equation}
\label{inverse_game}
\begin{aligned}
        \mathop{\max}\limits_{\theta, \capbstate, \capbcontrol}  \hspace{0.4cm} & {p(\capbobservation \given \capbstate, \capbcontrol)} & \\
        \text{s.t.} \hspace{0.4cm} & (\capbstate, \capbcontrol) \hspace{0.1cm} \text{is a} \hspace{0.1cm} \text{GNE} \hspace{0.1cm} \text{of} \hspace{0.1cm} \game(\theta) \\
\end{aligned}
\end{equation}

where~$p(\capbobservation\given\capbstate, \capbcontrol)$ denotes the likelihood of observations~$\capbobservation := (\bobservation_1, ..., \bobservation_T)$ given the estimated game trajectory~$(\capbstate, \capbcontrol)$ induced by parameters~$\theta$.
This formulation yields an \emph{\ac{mpec}}~\cite{luo1996mathematical}, where the outer problem is an estimation problem while the inner problem involves solving a dynamic game.
\revised{
When the observed game includes inequality constraints, the resulting inverse problem necessarily
contains complementarity constraints and only few tools are available to solve the resulting problem.
In the next section, we show how to transform \cref{inverse_game} into an unconstrained problem by making the inner game differentiable, which also enables combination with other differentiable components.
}

\textbf{Running example:} We assign the tracker (player~1) to be the ego-agent and parameterize the game with the goal position of the target robot $\theta^2 = p_{\text{goal}}^2$.
That is, the tracker does not know the target agent's goal and tries to infer this parameter from position observations.
To ensure that \cref{inverse_game} remains tractable, the ego-agent maintains only a fixed-length buffer of observed opponent's positions.
Note that solving the inverse game requires solving games rather than optimal control problems at the inner level to account for the noncooperative nature of observed interactions, which is different from \ac{ioc} even in the $2$-player case. 
We employ a Gaussian observation model, which we represent with an equivalent negative log-likelihood objective $\| \capbobservation - r(\capbstate, \capbcontrol) \|_2^2$ in \cref{inverse_game}, where $r(\capbstate, \capbcontrol)$ maps~$(\capbstate, \capbcontrol)$ to the corresponding sequence of expected positions.

\section{Adaptive Model-Predictive Game Play}
\label{sec:approach}
We wish to solve the problem of \ac{mpgp} from an ego-centric perspective, i.e., without prior knowledge of other players' objectives.
To this end, we present an adaptive model-predictive game solver that combines the tools of \cref{sec:preliminaries}:
first, we perform \ac{mle} of unknown objectives by solving an \emph{inverse game} (\cref{sec:inverse-game});
then, we solve a \emph{forward game} using this estimate to recover a strategic motion plan (\cref{sec:forward-game}).

\subsection{Forward Games as \acp{mcp}}
\label{sec:gnep-to-mcp}

We first discuss the conversion of the \ac{gnep} in \cref{forward_game}  to an equivalent \ac{mcp}.
There are three main advantages of taking this view.
First, there exists a wide range of off-the-shelf solvers for this problem class \cite{billups1997comparison}.
Furthermore, \ac{mcp} solvers directly recover strategies for all players \textit{simultaneously}. %
Finally, this formulation makes it easier to reason about derivatives of the solution w.r.t. to problem  data. %
As we shall discuss in~\cref{sec:adaptive-mpgp}, this derivative information can be leveraged to solve the inverse game problem of~\cref{inverse_game}.

In order to solve the \ac{gnep} presented in~\cref{forward_game} we derive its first-order necessary conditions.
We collect all equality constraints for player~$i$ in~\cref{forward_game} into a vector-valued function $h^i(\capstate^i, \capcontrol^i; \hat{\state}^i_1)$, introduce Lagrange multipliers~$\mu^i$,~${^p\lambda^i}$ and~${^s\lambda}$ for constraints~$h^i(\capstate^i, \capcontrol^i; \hat{\state}^i_1)$, ${^pg^i(\capstate^i,\capcontrol^i)}$, and ${^sg(\capbstate, \capbcontrol)}$ and write the Lagrangian for player $i$ as
~\begin{align}\label{eq:game-lagrangian}
&    \mathcal{L}^i(\capbstate, \capbcontrol, \mu^i, {^p\lambda^i}, {^s\lambda}; \theta) = \cost^i(\capbstate, \capbcontrol;\theta^i) \\
&+ \mu^{i\top}h^i(\capstate^i, \capcontrol^i; \hat{\state}^i_1) - {^{s}\lambda^{\top}}{^{s}g(\capbstate, \capbcontrol)} - {^{p}\lambda^{i\top}}{^{p}g^i(\capstate^i, \capcontrol^i)}.\nonumber
\end{align}
\revisedInline{Note that we share the multipliers associated with shared constraints between the
  players to encode equal constraint satisfaction responsibility~\cite{kulkarni2012variational}.}
Under mild regularity conditions, e.g., \ac{licq}, a solution of \cref{forward_game} must satisfy the following joint \ac{kkt} conditions:
\begin{equation}
\label{joint_kkt}
\begin{aligned}
\forall i \in [\numplayers]
\left \{
    \begin{aligned}
    \nabla_{(\capstate^i, \capcontrol^i)}\mathcal{L}^i(\capbstate, \capbcontrol, \mu^i, {^p\lambda^i}, {^s\lambda}; \theta) = 0 \\
    0 \leq {^{p}g^i(\capstate^i, \capcontrol^i)} \perp {^p\lambda^i} \geq 0\\
    \end{aligned}
\right.\\
    h(\capbstate, \capbcontrol; \hat{\bstate}_1) = 0\\
    0 \leq {^{s}g(\capbstate, \capbcontrol)} \perp {^s\lambda} \geq 0,
\end{aligned}
\end{equation}
where, for brevity, we denote by $h(\capbstate, \capbcontrol; \hat{\bstate}_1)$ the aggregation of all equality constraints.
If the second directional derivative of the Lagrangian is positive along all feasible directions at a solution of \cref{joint_kkt}---a condition that can be checked a posteriori---this point is also a solution of the original game.
In this work, we solve trajectory games by viewing their \ac{kkt} conditions through the lens of \acp{mcp} \cite[Section 1.4.2]{mcp_ref}.

\definition A \acrfull{mcp} is defined by the following problem data: a function $F(z): \R^d \mapsto \R^d$, lower bounds $\ell_j \in \R \cup \{-\infty\}$ and upper bounds $u_j \in \R \cup \{\infty\}$, each for~$j \in [d]$.
The solution of an \ac{mcp} is a vector $z^* \in \R^n$, such that for each element with index $j \in [  d]$ one of the following equations holds:
~\begin{subequations}
\begin{align}
    z^*_j = \ell_j, F_j(z^*) \geq 0 \label{eq: s1}\\
    \ell_j < z^*_j < u_j, F_j(z^*) = 0 \label{eq: s2}\\
    z^*_j = u_j, F_j(z^*) \leq 0 \label{eq: s3}.
\end{align}
\end{subequations}
The parameterized KKT system of \cref{joint_kkt} can be expressed as a \textit{parameterized family} of \acp{mcp} with decision variables corresponding to the primal and dual variables of \cref{joint_kkt},
\begin{equation*}%
    z =
    \left[
    \capbstate^\top, \capbcontrol^\top,
    \boldsymbol{\mu}^\top,
    {^{p}\lambda^{1\top}},
    \hdots,
    {^{p}\lambda^{\numplayers \top}},
    {^{s}\lambda^\top}
    \right]^\top,
\end{equation*}
\noindent and problem data
\begin{equation}
\label{game as mcp}
{\small
\begin{aligned}
F(z; \theta) &=\begin{bmatrix}
\nabla_{(\capstate^1, \capcontrol^1)}\mathcal{L}^i\\
\vdots\\
\nabla_{(\capstate^{\numplayers}, \capcontrol^{\numplayers})}\mathcal{L}^{\numplayers}\\
h\\
^{p}g^1\\
\vdots\\
^{p}g^{\numplayers}\\
^{s}g
\end{bmatrix}, &
\ell &= \begin{bmatrix}
- \infty\\
\vdots\\
- \infty\\
- \infty\\
0\\
\vdots\\
0\\
0
\end{bmatrix}, &
u &= \begin{bmatrix}
\infty\\
\vdots\\
\infty\\
\infty\\
\infty\\
\vdots\\
\infty\\
\infty
\end{bmatrix},
\end{aligned}
}
\end{equation}
where, by slight abuse of notation, we overload~$F$ to be parametrized by~$\theta$ via $\mathcal{L}^i$ and use $\infty$ to denote elements for which upper or lower bounds are dropped.

\subsection{Differentiation of an \ac{mcp} solver}
\label{differentiability}

An \ac{mcp} solver may be viewed as a function, mapping problem data to a solution vector.
Taking this perspective, for a parameterized family of \acp{mcp} as in \cref{game as mcp}, we wish to compute the function's derivatives to answer the following question:
How does the solution $z^*$ respond to local changes of the problem parameters $\theta$?

\subsubsection{The Nominal Case}
Let $\pmcp(\theta) := (F(\cdot; \theta), \ell, u)$ denote an \ac{mcp} parameterized by $\theta\in\R^p$ and let $z^*\in\R^n$ denote a solution of that \ac{mcp},
which is implicitly a function of $\theta$.
For this nominal case, we consider only solutions at which \emph{strict complementarity} holds. %
We shall relax this assumption later.
If $F$ is smooth, i.e.,~$F(\cdot; \theta), F(z^*; \cdot) \in C^1$, we can recover the Jacobian matrix~$\nabla_\theta z^*~=~\left(\frac{\partial z^*_j}{\partial \theta_k}\right) \in \R^{n\times p}$ by distinguishing two possible cases.
For brevity, below, gradients are understood to be evaluated at $z^*$ and $\theta$.

\paragraph{Active bounds}

Consider first the elements $z^*_j$ that are either at their lower or upper bound, i.e., $z^*_j$ satisfies \cref{eq: s1} or \cref{eq: s3}. %
Since strict complementarity holds at the solution, $F_j(z^*; \theta)$ must be bounded away from zero with a finite margin.
Hence, the smoothness of $F$ guarantees that a local perturbation of $\theta$ will retain the sign of $F_j(z^*; \theta)$.
As a result,~$z^*_j$  remains at its bound and, locally, is identically zero.
Let {\small $\tilde{\mc{I}} := \{k \in [n] \mid z_k^* = \ell_k \lor z_k^* = u_k\}$} denote the index set of all elements matching this condition and $\tilde{z}^* := [z^*]_{\tilde{\mc{I}}}$ denote the solution vector reduced to that set.
Trivially, then, the Jacobian of this vector vanishes,~i.e.,~$\nabla_\theta \tilde{z}^* = 0$.

\paragraph{Inactive bounds}

The second case comprises elements that are strictly between the bounds,~i.e.,~$z_j^*$ satisfying~\cref{eq: s2}.
In this case, under mild assumptions on $F$, for any local perturbation of~$\theta$ there exists a perturbed solution such that $F$ remains at its root.
Therefore, the gradient~$\nabla_\theta z_j^*$ for these elements is generally non-zero, and we can compute it via the \ac{ift}.
Let $\bar{\mc{I}} := \{k \in [n] \given F_k(z^*; \theta) = 0, \ell_k < z^*_k < u_k\}$ be the index set of all elements satisfying case~(b) and let
~\begin{equation}
\begin{aligned}
\bar{z}^* &:= [z^*]_{\bar{\mc{I}}}, & \bar{F}(z^*, \theta)  &:= [F(z^*; \theta)]_{\bar{\mc{I}}}
\end{aligned}
\end{equation}
denote the solution vector and its complement reduced to said index set.
By the \ac{ift}, the relationship between parameters~$\theta$ and solution~$z^*(\theta)$ is characterized by the stationarity of $\bar{F}$:
\begin{multline}\label{eq:f-stationary}
0 = \nabla_\theta\left[\bar{F}(z^*(\theta), \theta)\right] = \\
\nabla_\theta\bar{F} +
(\nabla_{\bar{z}^*}\bar{F})(\nabla_\theta\bar{z}^*) +
(\nabla_{\tilde{z}^*}\bar{F})\underbrace{(\nabla_\theta\tilde{z}^*)}_{\equiv 0}
\end{multline}
Note that, as per the discussion in case (a), the last term in this equation is identically zero.
Hence, if the Jacobian $\nabla_{\bar{z}^*}\bar{F}$ is invertible, we recover the derivatives as the unique solution of the above system of equations,
\begin{equation}\label{eq:ift}
\nabla_\theta\bar{z}^* =
     -\left(\nabla_{\bar{z}^*}\bar{F}\right)^{-1}(\nabla_\theta\bar{F}).
\end{equation}
Note that \cref{eq:f-stationary} may not always have a unique solution, in which case \cref{eq:ift} cannot be evaluated.
We discuss practical considerations for this special case below.

\subsubsection{Remarks on Special Cases and Practical Realization}

The above derivation of gradients for the nominal case involves several assumptions on the structure of the problem.
\revised{
We discuss considerations to improve numerical robustness for practical realization
of this approach below. We note that both special cases discussed hereafter are rare in practice.
In fact, across 100 simulations of the running example with varying initial states and objectives,
neither of them occurred.
}

\paragraph{Weak Complementarity}

The nominal case discussed above assumes strict complementarity at the solution.
If this assumption does not hold, the derivative of the \ac{mcp} is not defined.
Nevertheless, we can still compute subderivatives at~$\theta$.
Let the set of all indices for which this condition holds be denoted by~$\hat{\mc{I}} := \{k \in [n] \given F_k(z^*; \theta) = 0 \land z_k^* \in \{\ell_k, u_k\}\}$.
Then by selecting a subset of $\hat{\mc{I}}$ and including it in $\bar{\mc{I}}$ for evaluation of \cref{eq:ift}, we recover a subderivative.

\paragraph{Invertibility}
The evaluation \cref{eq:ift} requires invertibility of $\nabla_{\bar{z}^*}\bar{F}$.
To this end, we compute the least-squares solution of \cref{eq:f-stationary} rather than explicitly inverting $\nabla_{\bar{z}}\bar{F}$.%

\subsection{Model-Predictive Game Play with Gradient Descent}
\label{sec:adaptive-mpgp}
Finally, we present our pipeline for adaptive game-play against opponents with unknown objectives.
Our adaptive~\ac{mpgp} scheme is summarized in~\cref{mpgp}.
At each time step, we first update our estimate of the parameters by approximating the inverse game in~\cref{inverse_game} via gradient descent.
\revisedInline{To obtain an unconstrained optimization problem, we substitute the constraints in~\cref{inverse_game} with our differentiable game solver.}
Following the discussion of \cref{game as mcp}, we denote by $z^*(\theta)$ the solution of the \ac{mcp} formulation of the game parameterized by $\theta$.
Furthermore, by slight abuse of notation, we overload $\capbstate(z^*), \capbcontrol(z^*)$ to denote functions that extract the state and input vectors from $z^*$. %
Then, the inverse game of~\cref{inverse_game} can be written as unconstrained optimization,
\begin{equation}
\label{inverse_game_substituted}
\begin{aligned}
        \mathop{\max}\limits_{\theta}  \hspace{0.4cm} & {p(\capbobservation \given \capbstate(z^*(\theta)), \capbcontrol(z^*(\theta)))}.%
\end{aligned}
\end{equation}

Online, we approximate solutions of this problem by taking gradient descent steps on the negative logarithm of this objective, with
gradients computed by chain rule,
\begin{equation}\label{eq:full-chainrule}
\begin{split}
    \nabla_\theta\left[p(\capbobservation \mid \capbstate(z^*(\theta)), \capbcontrol(z^*(\theta))\right] = \\
    (\nabla_\capbstate p)(\nabla_{z^*}\capbstate)(\nabla_\theta z^*) +
    (\nabla_\capbcontrol p)(\nabla_{z^*}\capbcontrol)(\nabla_\theta z^*).
\end{split}
\end{equation}
Here, the only non-trivial term is~$\nabla_\theta z^*$, whose computation we discussed in \cref{differentiability}.
To reduce the computational cost, we warm-start using the estimate of the previous time step and terminate early if a maximum number of steps is reached.
Then, we solve a forward game parametrized by the estimated~$\Tilde{\theta}$ to compute control commands.
We execute the first control input for the ego agent and repeat the procedure.

\begin{algorithm}[t]
\caption{Adaptive \ac{mpgp}}
\label{mpgp}
{\scriptsize
\DontPrintSemicolon
\textbf{Hyper-parameters:} stopping tolerance: $\mathrm{stop\_tol}$, learning rate: $\mathrm{lr}$\\
\textbf{Input:} initial $\Tilde{\theta}$, current observation buffer $\capbobservation$, 
new observation $\bobservation$\\
$\capbobservation \gets$ updateBuffer$(\capbobservation, \bobservation)$\\
\tcc{inverse game approximation}
\While{$\mathrm{not~stop\_tol~and~not~max\_steps\_reached}$}{
    $(z^*, \nabla_\theta z^*) \gets \operatorname{solveDiffMCP}(\tilde{\theta})$\Comment{sec.~\ref{differentiability}}\\
    $\nabla_\theta p  \gets \operatorname{composeGradient}(z^*, \nabla_\theta z^*, \capbobservation)$\Comment{eq.~(\ref{eq:full-chainrule})}\\
    $\Tilde{\theta} \gets \Tilde{\theta} - \nabla_\theta p \cdot \mathrm{lr}$\\
}
$z^* \gets \operatorname{solveMCP}(\Tilde{\theta})$ \Comment{forward game, eq.~(\ref{game as mcp}})\\
$\operatorname{applyFirstEgoInput}(z^*)$\\
\textbf{return} $\tilde{\theta}, \capbobservation$
}
\end{algorithm}

\section{Experiments}
\label{sec:expriments}

To evaluate our method, we compare against two baselines in Monte Carlo studies of simulated
interaction.
Beyond these quantitative results, we showcase our method deployed on Jackal ground robots in two hardware experiments.

\revised{The experiments below are designed to support the key claims that our method
(\romannumeral 1)~outperforms both game-theoretic and non-game-theoretic baselines in highly interactive scenarios,
(\romannumeral 2)~can be combined with other differentiable components such as \acp{nn}, and
(\romannumeral 3)~is sufficiently fast and robust for real-time planning on a hardware platform.}
A supplementary video of qualitative results can be found at \href{https://xinjie-liu.github.io/projects/game}{https://xinjie-liu.github.io/projects/game}.
Upon publication of this manuscript, the code for our method and experiments will be available at the same link.

\vspace{-1.5em}
\revised{
\subsection{Experiment Setup}\label{sec:experiment-setup}
\subsubsection{Scenarios}\label{sec:scenarios}
We evaluate our method in two scenarios.
\paragraph{2-player running example}
To test the inference accuracy and convergence of our method in an intuitive setting, we first
consider the 2-player running example.
For evaluation in simulation, we sample the opponent's
intent---\ie, their unknown goal position in \cref{cost structure}--- uniformly from the environment.
Partial observations comprise the position of each agent.

\paragraph{Ramp merging}
To demonstrate the scalability of our approach and support the claim that our solver outperforms the
baselines in highly interactive settings, we also test our method on a ramp merging scenario with
varying numbers of players.
This experiment is inspired by the setup used in~\cite{cleac2022algames} and is schematically
visualized in~\cref{fig:motivation}.
We model each player's dynamics by a discrete-time kinematic bicycle with the state comprising position, velocity and orientation, \ie, $x_t^i = (p_{x,t}^i, p_{y,t}^i, v_{t}^i, \psi_{t}^i)$, and controls comprising acceleration and steering angle, \ie, $u_t^i = (a_t^i, \phi)$.
We capture their individual behavior by a cost function that penalizes deviation from a reference travel velocity and target lane; \ie,
$\theta^i = (v_\textrm{ref}^i, p_\textrm{y,lane}^i)$.
We add constraints for lane boundaries, for limits on speed, steering, and acceleration, for the traffic light, and for collision avoidance.
To encourage rich interaction in simulation, we sample each agent's initial state by sampling their speed and longitudinal positions
uniformly at random from the intervals from zero to maximum velocity $v_\mathrm{max}$ and four times the vehicle length ${l_\mathrm{car}}$, respectively.
The ego-agent always starts on the ramp and all agents are initially aligned with their current lane.
Finally, we sample each opponent's intent from the uniform distribution over the two lane centers and the target speed interval $[0.4v_\mathrm{max}, v_\mathrm{max}]$.
Partial observations comprise the position and orientation of each agent.
}

\subsubsection{Baselines}\label{sec:baselines}
We consider the following \revised{three} baselines. %

\paragraph{KKT-Constrained Solver}
In contrast to our method, the solver by Peters~\etalcite{peters2021inferring} has no support for either private or shared inequality constraints.
Consequently, this baseline can be viewed as solving a simplified version of the problem in \cref{inverse_game} where the inequality constraints associated with the inner-level \ac{gnep} are dropped.
Nonetheless, we still use a cubic penalty term as in \cref{cost structure} to encode soft collision avoidance.
Furthermore, for fair comparison, we only use the baseline to \emph{estimate} the objectives but compute control commands from a \ac{gnep} considering all constraints.

\paragraph{\ac{mpc} with Constant-Velocity Predictions}
This baseline assumes that opponents move with constant velocity as observed at the latest time step.
We use this baseline as a representative method for predictive planning approaches that do not explicitly model interaction.
\revisedInline{
\paragraph{Heuristic Estimation MPGP}
To highlight the importance of online intent inference, for the ramp merging evaluation, we also compare against a game-theoretic baseline that assumes a fixed intent for all opponents. This fixed intent is recovered by taking each agent's initial lane and velocity as a heuristic preference estimate.
}

To ensure a fair comparison, we use the same \ac{mcp} backend \cite{dirkse1995path} to solve all \ac{gnep}s and optimization problems \revised{with a default convergence tolerance of $1e^{-6}$.}
Furthermore, all planners utilize the same planning horizon and history buffer size of~$10$ time steps with a time-discretization of~$\SI{0.1}{\second}$. 
\revised{For the iterative \ac{mle} solve procedure in the $2$-player running example and the ramp merging scenario, we employ a learning rate of $2e^{-2}$ for objective parameters and $1e^{-3}$ for initial states.
We terminate \acl{mle} iteration when the norm of the parameter update step is smaller than
$1e^{-4}$, or after a maximum of $30$ steps.}
Finally, opponent behavior is generated by solving a separate ground-truth game whose parameters are hidden from the ego-agent.

\subsection{Simulation Results}
\revisedInline{
To compare the performance of our method to the baselines described in~\cref{sec:baselines}, we
conduct a Monte Carlo study for the two scenarios described in~\cref{sec:scenarios}.}
\subsubsection{$2$-Player Running Example}
\begin{figure}
\bigskip
    \centering
    \includegraphics[scale=0.16]{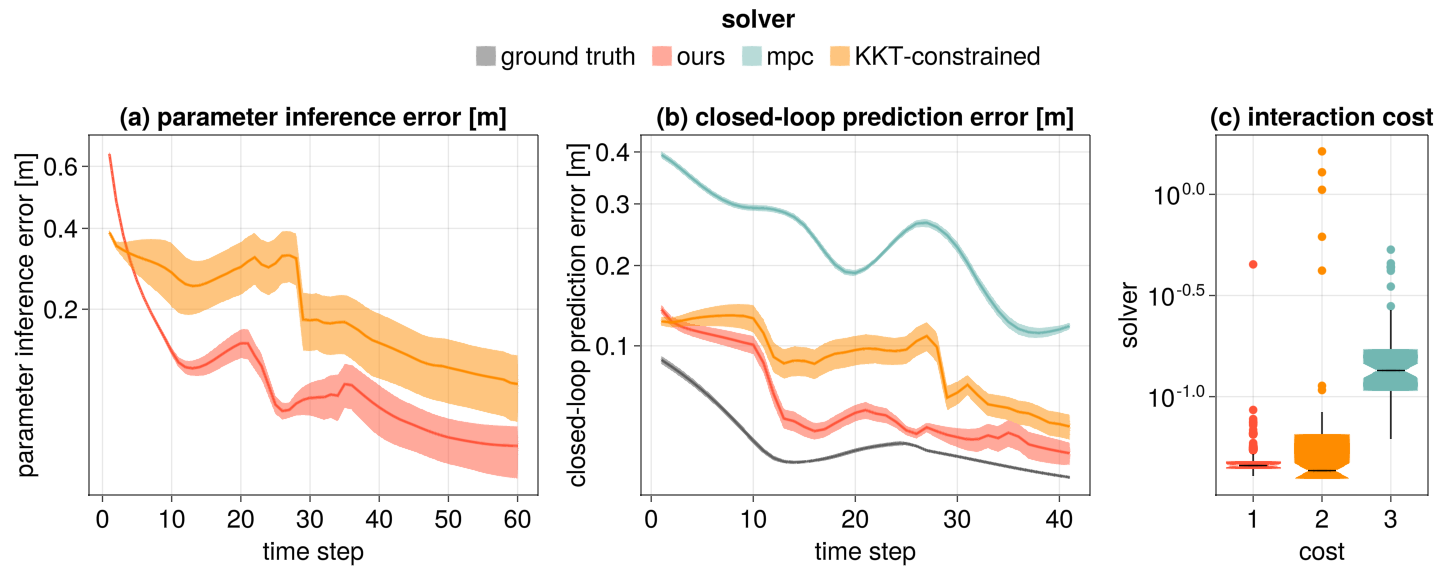}
    \caption{Monte Carlo study for the $2$-player tracking game for~$100$ trials. Solid lines and ribbons in (a) and (b) indicate the mean and \acl{sem}.
    Cost distributions in \revised{(c)} are normalized by subtracting ground truth costs.}
    \label{fig:running example}
\end{figure}

\Cref{fig:running example} summarizes the results for the 2-player running example.
For this evaluation, we filter out any runs for which a solver resulted in a collision.
For our solver, the KKT-constrained baseline, and the \ac{mpc} baseline this amounts to $2, 2$ and $13$ out of $100$ episodes, respectively.

Figures~\ref{fig:running example}(a-b) show the prediction error of the goal position and opponent's trajectory, each of which is measured by~$\ell^2$-norm.
Since the \ac{mpc} baseline does not explicitly reason about costs of others, we do not report parameter inference error for it in~\cref{fig:running example}a.
As evident from this visualization, both game-theoretic methods give relatively accurate parameter estimates and trajectory predictions.
Among these methods, our solver converges more quickly and consistently yields a lower error.
By contrast, \ac{mpc} gives inferior prediction performance with reduced errors only in trivial cases, when the target robot is already at the goal.
\Cref{fig:running example}c shows the distribution of costs incurred by the ego-agent for the same set of experiments.
Again, game-theoretic methods yield better performance and our method outperforms the baselines with more consistent and robust behaviors, indicated by fewer outliers and lower variance in performance.

\subsubsection{Ramp Merging}\label{sec:ramp-merging-results}
\input{highwayTable}
\begin{figure}
    \centering
    \begin{subfigure}{0.9\linewidth}
        \centering
        \includegraphics[width=1\linewidth]{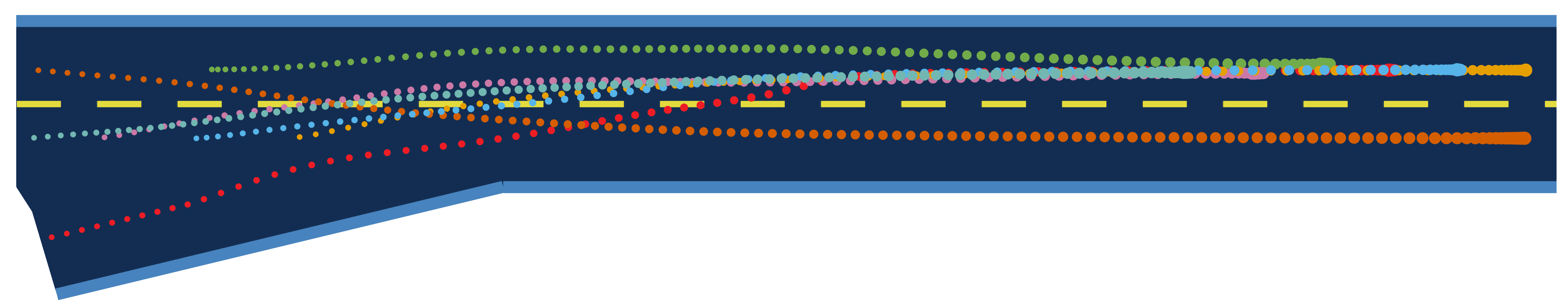}
        \vspace{-25pt}
        \caption{\revised{Qualitative performance.}}
        \vspace{10pt}
    \end{subfigure}
    \begin{subfigure}{\linewidth}
    \centering
    {
    \tiny\begin{tabular}{c|c|c|c|c|c|c}
    \toprule
    \textbf{\makecell{Ego \\ cost}} & \textbf{\makecell{Opp.\\ cost}} & \textbf{\makecell{Coll.}} &
    \textbf{\makecell{Inf.}} & \textbf{\makecell{Traj.\\ err. [m]}} & \textbf{\makecell{Param.\\
    err.}} & \textbf{\makecell{Time [2]}} \\
    \midrule
    \makecell{2.19 \\$\pm$ 1.21} & \makecell{0.17 \\$\pm$ 0.07} & \makecell{3}  & \makecell{5}   & \makecell{2.34\\  $\pm$ 0.08} & \makecell{0.91\\ $\pm$ 0.08} & \makecell{0.274\\ $\pm$ 0.01} \\
    \bottomrule
    \end{tabular}}
    \caption{\revised{Quantitative performance.}}
    \end{subfigure}
    \caption{
      Performance of our solver in combination with an \ac{nn} for 100 trials of the 7-player ramp merging scenario.
    }
    \label{highway-traj}
\end{figure}
\revisedInline{
\Cref{tab:ramp merging} summarizes the results of for the simulated ramp-merging scenario for 3, 5, and 7 players.
\paragraph{Task Performance}
To quantify the task performance, we report costs as an indicator for interaction efficiency, the number of collisions as a measure of safety, number of infeasible solves as an indicator of robustness, and trajectory and parameter error as a measure of inference accuracy.}
\revised{
On a high level, we observe that the game-theoretic methods generally outperform the other baselines; especially for the settings with higher traffic density.
While \ac{mpc} achieves high efficiency (ego-cost) in the 3-player case, it collides significantly more often than the other methods across all settings.
Among the game-theoretic approaches, we observe that online inference of opponent intents---as performed by our method and the KKT-constrained baseline---yields better performance than a game that uses a heuristic estimate of the intents.
Within the inference-based game solvers, a Manning-Whitney U-test reveals that, across all settings, both methods achieve an ego-cost that is significantly lower than all other baselines but not significantly higher than solving the game with ground truth opponent intents.
Despite this tie in terms of interaction \emph{efficiency}, we observe a statistically significant improvement of our method over the KKT-constrained baseline in terms of \emph{safety}: in the highly interactive 7-player case, the KKT-constrained baseline collides seven times more often than our method.
This advantage is enabled by our method's ability to model inequality constraints within the inverse game.

\paragraph{Computation Time}
We also measure the computation time of each approach.
The inference-based game solvers have generally a higher runtime than the remaining methods due to the added complexity. Within the inference methods, our method is only marginally slower than the KKT-constrained
baseline, despite solving a more complex problem that includes inequality constraints.
The average number of \ac{mle} updates for our method was $11.0$, $19.2$, and $22.7$ for the 3, 5, and 7-player setting, respectively.
While our current implementation achieves real-time planning rates only for up to three players, we note that additional optimizations may further reduce the runtime of our approach.
Among such optimizations are low-level changes such as sharing memory between \ac{mle} updates as well as algorithmic changes to perform intent inference asynchronously at an update rate lower than the control rate.
We briefly explore another algorithmic optimization in the next section.
}

\subsubsection{Combination with an \ac{nn}}\label{sec:nn}
To support the claim that our method can be combined with other differentiable modules, we
demonstrate the integration with an \ac{nn}.
For this proof of concept, we use a two-layer feed-forward \ac{nn}, which takes the buffer of recent partial state observations as input and predicts other players' objectives.
Training of this module is enabled by propagating the gradient of the observation likelihood loss of \cref{inverse_game_substituted} through the differentiable game solver to the parameters of the \ac{nn}.
Online, we use the network's prediction as an initial guess to reduce the number gradient steps.
\revised{
As summarized in \cref{highway-traj}, this combination reduces the computation time by more than
60\% while incurring only a marginal loss in performance.
}

\subsection{Hardware Experiments}

\begin{figure}
\medskip
	\centering
	\begin{subfigure}{0.45\linewidth}
		\centering
		\includegraphics[width=1\linewidth]{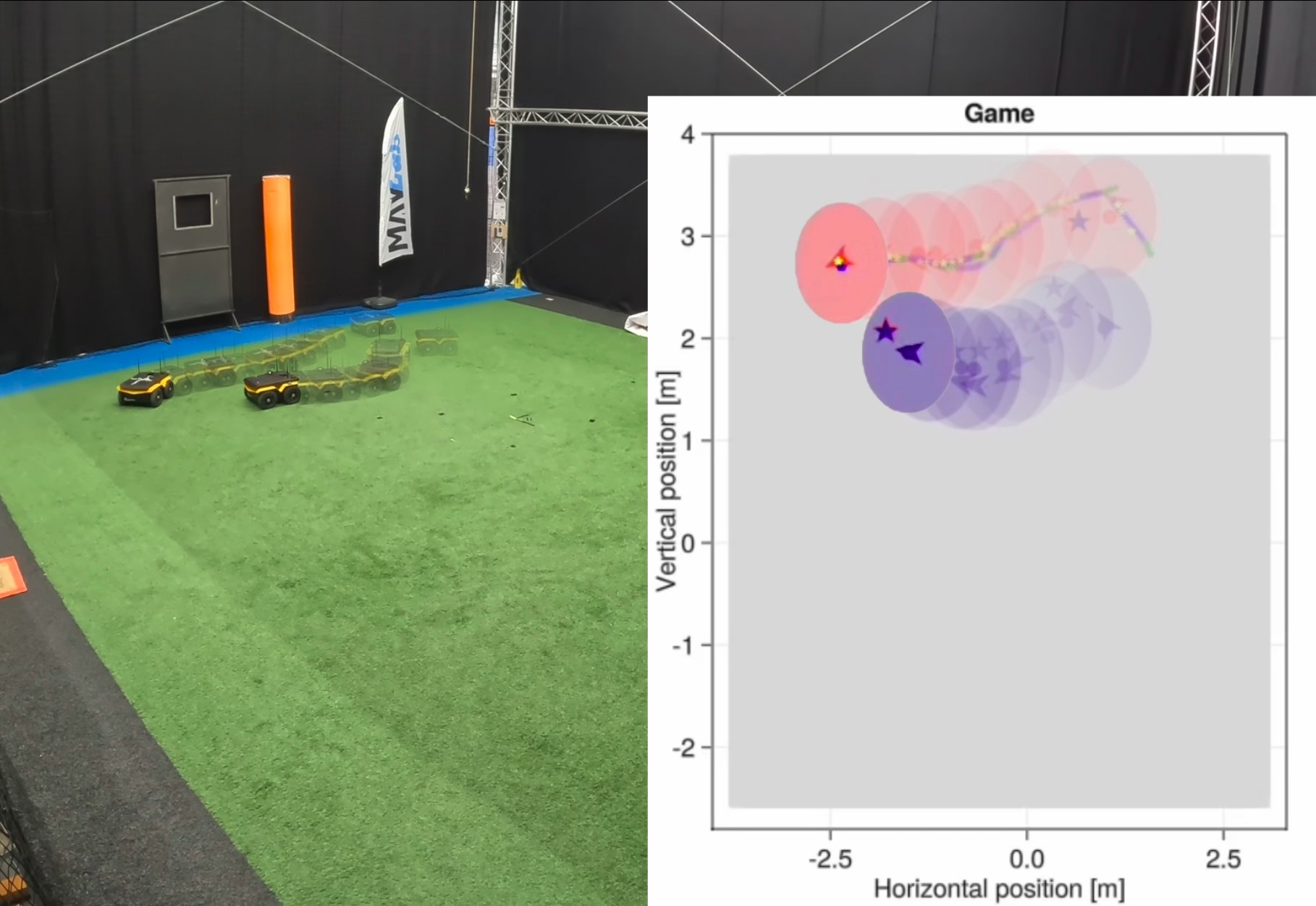}
		\caption{}
		\label{jackals}
	\end{subfigure}
	\begin{subfigure}{0.4\linewidth}
		\centering
		\includegraphics[width=1\linewidth]{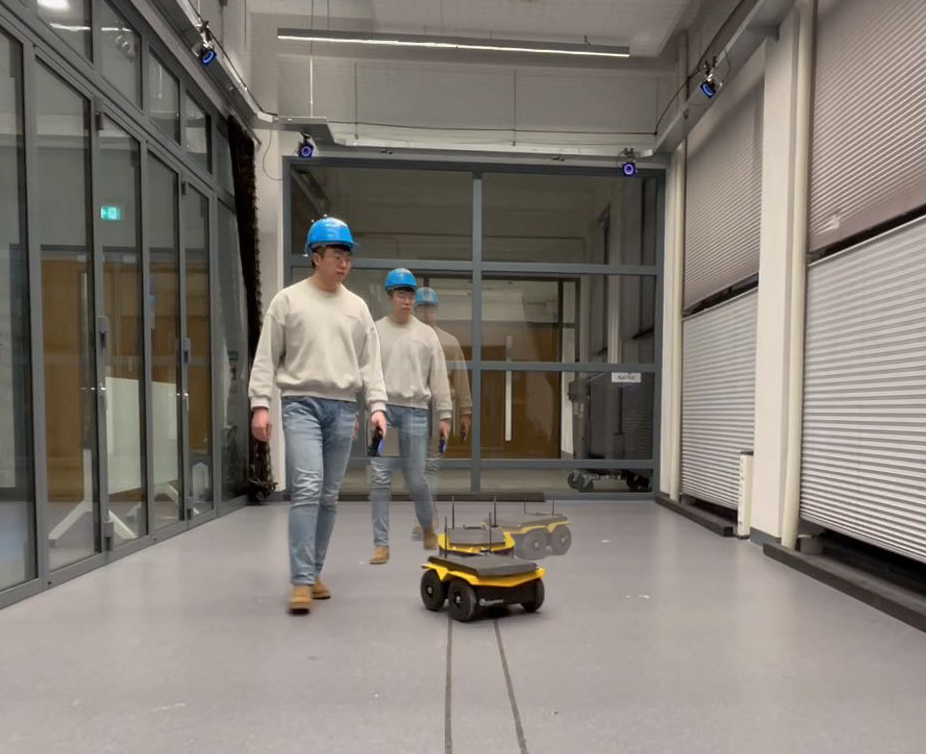}
		\caption{}
		\label{jackal_hri}
	\end{subfigure}
	\caption{Time lapse of the running-example in which a Jackal tracks (a) another Jackal and
    \revised{(b) a human}. Overlaid in (a) are the position of target robot (red) its true goal (red
    star), the tracker (blue), and its goal estimate (blue star).
	}
	\label{jackal}
\end{figure}

To support the claim that our method is sufficiently fast and robust for hardware deployment, we demonstrate the tracking game in the running example in~\cref{running example} with a Jackal ground robot tracking (\romannumeral 1) another Jackal robot (\cref{jackals}) and (\romannumeral 2) a human player (\cref{jackal_hri}), each with initially unknown goals.
Plans are computed online on a mobile i$7$ CPU.
\revised{
  We generate plans using the point mass dynamics with a velocity constraint of \SI{0.8}{\meter\per\second} and realize low-level control via the feedback controller of \cite{kanayama1990stable}.
}
A video of these hardware demonstrations is included in the supplementary material.
\revised{
In both experiments, we observe that our adaptive \ac{mpgp} planner enables the robot to infer the
unknown goal position to track the target while avoiding collisions.
The average computation time in both experiments was \SI{0.035}{\second}.
}

\section{Conclusion}

In this paper, we presented a model-predictive game solver that adapts strategic motion plans to initially unknown opponents' objectives.
The adaptivity of our approach is enabled by a differentiable trajectory game solver whose gradient signal is used for \ac{mle} of unknown game parameters.
As a result, our adaptive \ac{mpgp} planner allows for safe and efficient interaction with other strategic agents without assuming prior knowledge of their objectives or observations of full states.
We evaluated our method in two simulated interaction scenarios and demonstrated superior performance over a state-of-the-art game-theoretic planner and a non-interactive \ac{mpc} baseline.
Beyond that, we demonstrated the real-time planning capability and robustness of our approach in two hardware experiments.

\revised{
In this work, we have limited inference to parameters that appear in the objectives of other players.
Since the derivation of the gradient in~\cref{differentiability} can also handle other parameterizations of $F$---so long as they are smooth---future work may extend this framework to infer additional parameters of constraints or aspects of the observation model.
Furthermore, encouraged by the improved scalability when combining our method with learning modules such as \acp{nn}, we seek to extend this learning pipeline in the future.
One such extension would be to operate directly on raw sensor data, such as images, to exploit
additional visual cues for intent inference.
Another extension is to move beyond MLE-based point estimates to inference of potentially multi-modal distributions over opponent intents, which may be achieved by embedding our differentiable  method within a variational autoencoder.
Finally, our framework could be tested on large-scale datasets of real autonomous-driving
behavior.
}

\bibliographystyle{IEEEtran}
\bibliography{liu2022ral}

\end{document}

%% file: highwayTable.tex
\begin{table}[t]
\vspace{1em}
\centering
{\tiny\begin{tabular}{c|c|c|c|c|c|c|c|c}%
\toprule
\textbf{Set.} & \textbf{Method} & \textbf{\makecell{Ego \\ cost}} & \textbf{\makecell{Opp.\\ cost}} & \textbf{\makecell{Coll.}} & \textbf{\makecell{Inf.}} & \textbf{\makecell{Traj.\\ err. [m]}} & \textbf{\makecell{Param.\\ err.}} & \textbf{\makecell{Time [s]}} \\
\midrule
\multirow{8}{*}{\sw{3 player}}  & Ours      & \makecell{0.64 \\$\pm$ 0.36} & \makecell{0.06 \\$\pm$ 0.03} & \makecell{0}  & \makecell{0}   & \makecell{1.29\\  $\pm$ 0.05} & \makecell{0.41\\ $\pm$ 0.03} & \makecell{0.081\\ $\pm$ 0.002} \\ \cline{2-9}
                                & KKT-con   & \makecell{1.85 \\$\pm$ 1.21} & \makecell{0.05 \\$\pm$ 0.02} & \makecell{0}  & \makecell{1}   & \makecell{1.32\\  $\pm$ 0.06} & \makecell{2.39\\ $\pm$ 0.11} & \makecell{0.060\\ $\pm$ 0.002} \\ \cline{2-9}
                                & Heuristic & \makecell{6.73 \\$\pm$ 2.40} & \makecell{0.09 \\$\pm$ 0.07} & \makecell{0}  & \makecell{11}  & \makecell{7.89\\  $\pm$ 0.26} & \makecell{3.96\\ $\pm$ 0.13} & \makecell{0.008\\ $\pm$ 0.001} \\ \cline{2-9}
                                & MPC       & \makecell{1.50 \\$\pm$ 0.45} & \makecell{0.33 \\$\pm$ 0.07} & \makecell{28} & \makecell{218} & \makecell{2.40\\  $\pm$ 0.11} & \makecell{n/a}               & \makecell{0.009\\ $\pm$ 0.002} \\ \midrule
\multirow{8}{*}{\sw{5 player}}   & Ours      & \makecell{0.56 \\$\pm$ 0.43} & \makecell{0.16 \\$\pm$ 0.06} & \makecell{0}  & \makecell{2}   & \makecell{1.66\\  $\pm$ 0.07} & \makecell{0.47\\ $\pm$ 0.03} & \makecell{0.29 \\$\pm$ 0.02}   \\ \cline{2-9}
                                 & KKT-con   & \makecell{0.07 \\$\pm$ 0.32} & \makecell{0.06 \\$\pm$ 0.02} & \makecell{1}  & \makecell{4}   & \makecell{1.70\\  $\pm$ 0.06} & \makecell{2.15\\ $\pm$ 0.06} & \makecell{0.28 \\$\pm$ 0.02}   \\ \cline{2-9}
                                 & Heuristic & \makecell{2.06 \\$\pm$ 0.44} & \makecell{0.35 \\$\pm$ 0.10} & \makecell{5}  & \makecell{25}  & \makecell{8.05\\  $\pm$ 0.19} & \makecell{2.91\\ $\pm$ 0.07} & \makecell{0.015\\ $\pm$ 0.001} \\ \cline{2-9}
                                 & MPC       & \makecell{5.73 \\$\pm$ 2.91} & \makecell{0.42 \\$\pm$ 0.13} & \makecell{44} & \makecell{552} & \makecell{2.87\\  $\pm$ 0.13} & \makecell{n/a}               & \makecell{0.014\\ $\pm$ 0.002} \\ \midrule
\multirow{9}{*}{\sw{7 player}}   & Ours      & \makecell{1.60 \\$\pm$ 1.19} & \makecell{0.06 \\$\pm$ 0.02} & \makecell{1}  & \makecell{1}   & \makecell{1.89\\  $\pm$ 0.05} & \makecell{0.46\\ $\pm$ 0.02} & \makecell{0.68 \\$\pm$ 0.02}    \\ \cline{2-9}
                                 & KKT-con   & \makecell{3.11 \\$\pm$ 1.72} & \makecell{0.09 \\$\pm$ 0.04} & \makecell{7}  & \makecell{22}  & \makecell{2.01\\  $\pm$ 0.06} & \makecell{1.93\\ $\pm$ 0.03} & \makecell{0.63 \\$\pm$ 0.06}    \\ \cline{2-9}
                                 & Heuristic & \makecell{6.60 \\$\pm$ 1.67} & \makecell{0.27 \\$\pm$ 0.06} & \makecell{8}  & \makecell{8}   & \makecell{8.18\\  $\pm$ 0.15} & \makecell{2.44\\ $\pm$ 0.05} & \makecell{0.031\\ $\pm$ 0.002}  \\ \cline{2-9}%
                                 & MPC       & \makecell{8.41 \\$\pm$ 1.45} & \makecell{0.59 \\$\pm$ 0.09} & \makecell{43} & \makecell{848} & \makecell{3.07\\  $\pm$ 0.08} & \makecell{n/a}               & \makecell{0.0274\\ $\pm$ 0.004} \\

\bottomrule
\end{tabular}
}
\caption{\revised{Monte Carlo study for the ramp merging scenario depicted in~\cref{fig:motivation}
  with~$100$ trials for settings with 3, 5, and 7 players.}
  Except for collision and infeasible solve times, all metrics are reported by mean and \acrlong{sem}.
}
\label{tab:ramp merging}
\vspace{-0.3cm}
\end{table}